# OCR of historical printings with an application to building diachronic corpora: A case study using the RIDGES herbal corpus

Uwe Springmann
LMU Munich & Humboldt-Universität zu Berlin
uwe.springmann@hu-berlin.de

Anke Lüdeling
Humboldt-Universität zu Berlin
anke.luedeling@rz.hu-berlin.de

**Abstract**

This article describes the results of a case study that applies Neural Network-based Optical Character Recognition (OCR) to scanned images of books printed between 1487 and 1870 by training the OCR engine OCRopus (Breuel et al. 2013) on the RIDGES herbal text corpus (Odebrecht et al. 2017, in press). Training specific OCR models was possible because the necessary *ground truth* is available as error-corrected diplomatic transcriptions. The OCR results have been evaluated for accuracy against the ground truth of unseen test sets. Character and word accuracies (percentage of correctly recognized items) for the resulting machine-readable texts of individual documents range from 94% to more than 99% (character level) and from 76% to 97% (word level). This includes the earliest printed books, which were thought to be inaccessible by OCR methods until recently. Furthermore, OCR models trained on one part of the corpus consisting of books with different printing dates and different typesets *(mixed models)* have been tested for their predictive power on the books from the other part containing yet other fonts, mostly yielding character accuracies well above 90%. It therefore seems possible to construct generalized models trained on a range of fonts that can be applied to a wide variety of historical printings still giving good results. A moderate postcorrection effort of some pages will then enable the training of individual models with even better accuracies. Using this method, diachronic corpora including early printings can be constructed much faster and cheaper than by manual transcription. The OCR methods reported here open up the possibility of transforming our printed textual cultural heritage into electronic text by largely automatic means, which is a prerequisite for the mass conversion of scanned books.

# 1 Introduction

This paper describes a procedure for converting images of historical printings to electronic text with high accuracies, ranging on our test corpus from 94% to 99% (character accuracies) and 76% to 97% (word accuracies), by applying Neural Network-based Optical Character Recognition (OCR). The possibility to OCR historical printings with the same relative ease as more recent printings from the 20th century onward would be highly welcome to all researchers who have to deal with source materials from these periods. By historical printings we here summarily mean all documents from the beginning of modern printing in 1450 to the 19th century.

Optical Character Recognition (OCR) for modern printed texts using the Latin alphabet works very well and is often considered a solved problem (see Doermann and Tombre 2014, 256). Traditional OCR methods available in commercial and open source software products work as follows: During the OCR process, an image of a printed page is segmented into characters which are then compared to abstract feature sets describing prototypical characters learned previously from a set of trained fonts. The similarity of learned and recognized fonts, the clear separation of uniformly black characters and white spotless background, and modern standardized spelling of the printed words all contribute to excellent recognition results.

This is, however, not true for early printings because of non-standard typography, highly variable spelling preventing automatic lexical correction during the OCR process, and physical degradation of the pages due to aging and usage (see Figure 1 for two typical examples of pages from the RIDGES corpus, a diachronic corpus of herbal texts in German, which will be described in more detail in Section 3).

Historical typography alone poses a severe limit for the effectiveness of OCR: All available OCR engines have been extensively trained on modern fonts, but since historical fonts are very different from modern ones and the engines cannot be trained very well by end users on these historical fonts, training on modern fonts has only very limited value for the OCR of historical printings. Even for relatively late (i.e. 19th century) texts, OCR results (especially for the multitude of highly variable *broken* blackletter typefaces) from commercial OCR engines are often less than satisfactory (Piotrowski 2012; Strange et al. 2014). Even the more regular Antiqua typefaces (Roman glyph shapes) of historical printings often lead to character accuracies of only around 85% when they are OCRed with ABBYY Finereader[1], a leading commercial product (Reddy and Crane 2006; Springmann et al.

[1] http://https://www.abbyy.com/finereader/

2014). Tanner, Muñoz, and Ros (2009) report their experience from the British Libary's 19th Century Online Newspaper Archive[2] and state that character accuracies greater than 95% are "more usual for post-1900 and pre-1950's text and anything pre-1900 will be fortunate to exceed 85% accuracy[3]." Incunabula as the earliest texts (printed from 1450 to 1500) have been deemed to be completely unsuitable for OCR methods (Rydberg-Cox 2009)[4].

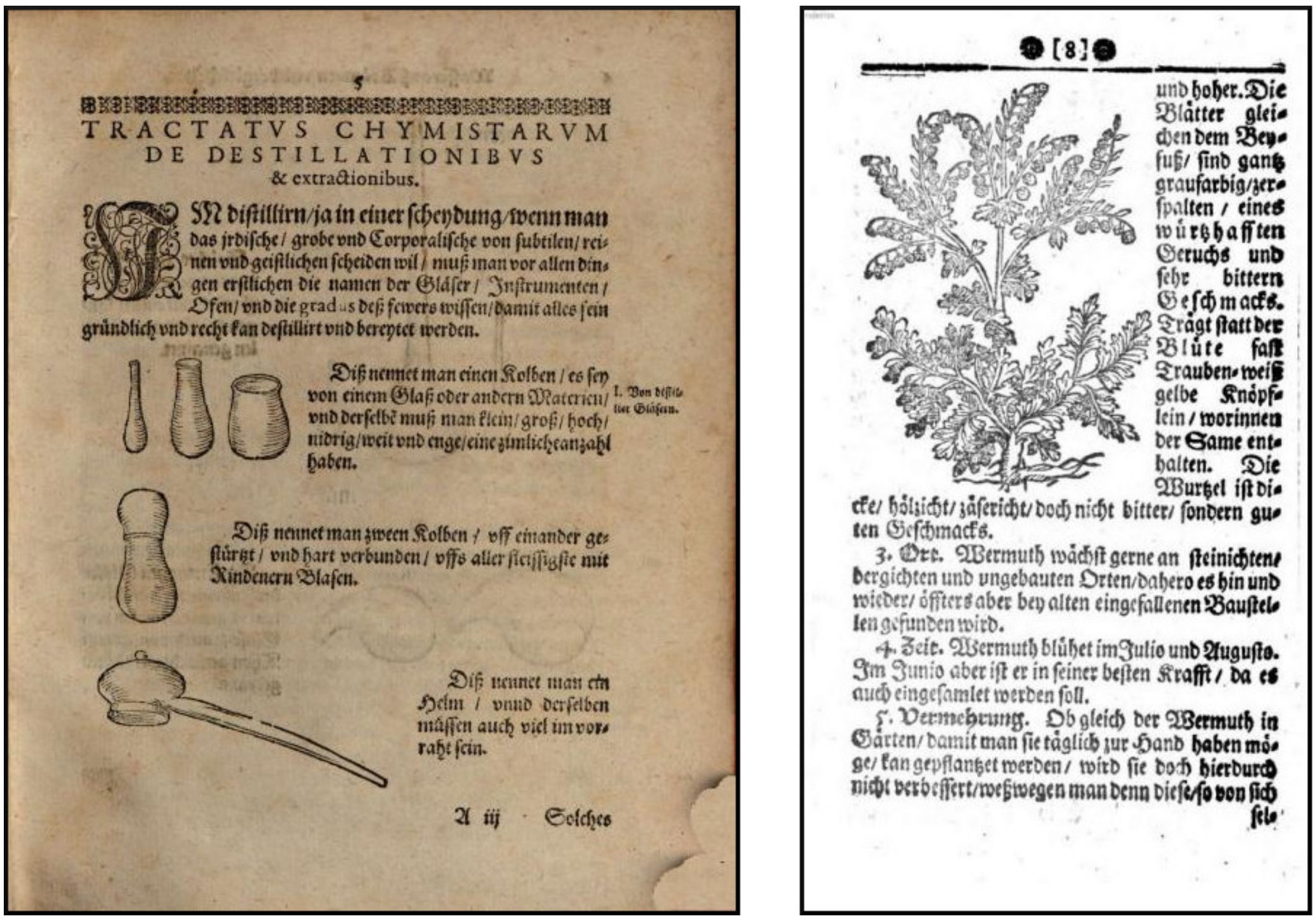
5

TRACTATVS CHYMISTARVM
DE DESTILLATIONIBVS
& extractionibus.

WJn distillirn/ja in einer scheydung/wenn man das jrdische/ grobe vnd Corporalische von subtilen/ reinen vnd geistlichen scheiden wil/ muß man vor allen dingen erstlichen die namen der Gläser/ Instrumenten/ Ofen/ vnd die gradus deß fewers wissen/ damit alles fein gründlich vnd recht kan destillirt vnd bereytet werden.

Diß nennet man einen Kolben/ es sey von einem Glaß oder andern Materien/ vnd derselbē muß man klein/ groß/ hoch/ nidrig/ weit vnd enge/ eine zimlicheanzahl haben. 1. Von destillier Gläsern.

Diß nennet man zween Kolben/ vff einander gestürtzt/ vnd hart verbunden/ vffs aller fleissigste mit Rindenern Blasen.

Diß nennet man ein Helm/ vnnd derselben müssen auch viel im vorraht sein.

A iij Solches

[8]

und hoher. Die Blätter gleichen dem Beyfuß/ sind gantz graufarbig/ zerspalten/ eines würtzhafften Geruchs und sehr bittern Geschmacks. Trägt statt der Blüte fast Trauben-weiß gelbe Knöpflein/ worinnen der Same enthalten. Die Wurtzel ist dicke/ hölzicht/ zäsericht/ doch nicht bitter/ sondern guten Geschmacks.

3. Ort. Wermuth wächst gerne an steinichten/ bergichten und ungebauten Orten/ dahero es hin und wieder/ öffters aber bey alten eingefallenen Baustellen gefunden wird.

4. Zeit. Wermuth blühet im Julio und Augusto. Im Junio aber ist er in seiner besten Krafft/ da es auch eingesamlet werden soll.

5. Vermehrung. Ob gleich der Wermuth in Gärten/ damit man sie täglich zur Hand haben möge/ kan gepflantzet werden/ wird sie doch hierdurch nicht verbessert/ weßwegen man denn diese/ so von sich sel-

Figure 1: Two page examples of early printings (left: Libavius 1603; right: Curioser Botanicus, 1675) illustrating barriers to OCR: text mixed with images, historical fonts, bleed-through from back page (left), bad contrast (right).

The constraint that OCR engines only give good results on a fixed set of pre-trained fonts can be lifted by the fully trainable open-source OCR engines Tesseract[5] and OCRopus.[6] There are two ways of training, namely training on synthetic data (images generated from existing electronic text and available computer fonts), or training on real data (scans and their transcriptions). The first method obviates the need to generate *ground truth* data by diplomatic transcriptions (a recording of glyph shapes as they appear on a page, with no or minimal editorial interpretation) which are needed to establish the link from glyph shapes to characters during the training process, and one also does not need to preprocess real images. As the complete training process can be automated, this training method is the preferred one whenever it is applicable. However, the multitude of historical fonts is not matched by existing computer fonts, and the irregularities of word spacings in real printings lead to inferior recognition results compared to training on real data (Springmann et al. 2014). OCR training for historical printings therefore has to rely on a training process using real data, which means that diplomatic transcriptions of printed images become a key resource. Unfortunately, diplomatically transcribed historical corpora which could serve as training data for historical fonts are not yet available in sufficient quantity. One problem is variability: Earlier typographies are more variable than modern ones because the process of designing, producing and distributing metal letter types (type founding) had not yet become a profession of its own, and early printers had to produce their own typesets leading to a large variety of historical fonts. It is therefore problematic to learn the associations between printed glyphs and the characters they represent from one printing and apply them to the next.

[2]http://www.bl.uk/reshelp/findhelprestype/news/newspdigproj/database/index.html
[3]http://www.dlib.org/dlib/july09/munoz/07munoz.html
[4]http://digitalhumanities.org/dhq/vol/3/1/000027/000027.html#p7
[5]https://github.com/tesseract-ocr/tesseract
[6]https://github.com/tmbdev/ocropy

The problem of applying OCR methods to historical printings is thus twofold: First one needs to train an individual model for a specific book with its specific typography. This can be achieved by transcribing some portion of the printed text, which usually requires linguistic knowledge. Second, even if this model works well for the book it has been trained on, it does not normally produce good OCR results for other books, even if their fonts look similar to the human eye. We need to overcome this *typography barrier* in order to use OCR methods effectively in the building of a historical electronic text corpus.

In the following we will describe our experiments addressing both problems. First we describe a procedure for training *individual models*, using a new recognition algorithm based upon recurrent neural nets as implemented in the open-source OCR engine OCRopus (Breuel et al. 2013). As training material we use the readily available scans from library digitization programs and the transcriptions from the RIDGES corpus. The individual model is trained on a single document with its specific fonts and word distances and therefore optimally adapted to this specific book. These models yield the best recognition results for the unseen pages of the books for which they have been trained, but unfortunately give mostly bad results on any other books. Then we explore the viability of training *mixed models* on a range of different typographies, pooling the training material from a variety of books, with the hope that these models are able to better generalize to other books that had no representation in the training pool. All models, even the individual ones, only get tested on test sets that have no overlap with the training material that was used in model training.

The remainder of this paper is organized as follows: The current state of the art in historical OCR is related in Section 2. As we will illustrate and evaluate our method on the RIDGES corpus we will briefly describe it in the following section. Section 4 details our procedure for training individual models for a subset of 20 books printed in broken (blackletter) typefaces, which were used extensively in German printing until well into the 20th century. In Section 5 we present the training results and discuss the need for building generalized (mixed) models. Section 6 reports on our experiments to construct mixed models and their performance on unseen books. In Section 7 we dicuss the significance of our results for the building of historical corpora and Section 8 concludes with a summary.

## 2 Related work

Work on historical OCR by other groups has mostly focused on Tesseract which is trainable on artificial images generated from computer fonts. Training on real data, however, has proven to be difficult, and lead to efforts to reconstruct the original historical font from cut-out glyphs. This has been done by both the Digital Libraries Team of the Poznań Supercomputing and Networking Center (Dudczak, Nowak, and Parkoła 2014) with their cutouts application[7] (proprietary) and EMOP's Franken+ tool[8] (open-source). The latter group has reported[9] that they were able to reach a character accuracy of about 86% on the ECCO document collection[10] and 68% on the EEBO collection.[11] Their OCR results suffer badly from scans of binarized microfilm images containing a lot of noise.

The Kallimachos project[12] at Würzburg University did have success with Franken+ to reach character accuracies over 90% for an incunable printing (Kirchner et al. 2016), but this method again relies on creating diplomatic transcriptions from scratch for each individual font. Ul-Hasan, Bukhari, and Dengel (2016) proposed a method to circumvent ground truth production by first training Tesseract on a historically reconstructed font. They then applied the resulting model to a specific book and used the recognized text as pseudo-ground truth to train OCRopus. The newly recognized text was again used as pseudo-ground truth for another round of OCRopus training, and after a few iterations they also achieved character accuracies above 93%, but their method shifts the effort to the manual (re)construction of the printed font. Training OCRopus with erroneous OCR recognition as pseudo-ground truth was also tried as one of several methods to improve OCR quality in Springmann, Fink, and Schulz (2016) but was always found to yield inferior results compared to training on even a small amount of real ground truth production at the order of 100 to 200 printed lines which consistently lead to character accuracies above 95%.

[7]https://confluence.man.poznan.pl/community/display/WLT/Cutouts+application
[8]http://emop.tamu.edu/outcomes/Franken-Plus
[9]http://emop.tamu.edu/final-report
[10]Eighteenth Century Collections Online, http://quod.lib.umich.edu/e/ecco/
[11]Early English Books Online, http://www.textcreationpartnership.org/tcp-eebo/
[12]http://kallimachos.de

A completely different approach was taken with the new Ocular OCR engine by Berg-Kirkpatrick, Durrett, and Klein (2013), Berg-Kirkpatrick and Klein (2014) which is able to convert printed to electronic text in a completely unsupervised manner (i.e., no ground truth is needed) employing a language, typesetting, inking, and noise model. This may be a viable alternative for training individual models with low manual effort, but it seems to be very resource-intensive and slow (transcribing 30 lines of text in 2.4 min according to Berg-Kirkpatrick and Klein 2014). Its results are better than (untrained) Tesseract and ABBYY, but it remains to be shown that they consistently reach character accuracies higher than 90%.

In summary, while there are other approaches to train individual OCR models for the recognition of historical documents, none have so far reported results as good as OCRopus when trained on real data (consistently over 94% character accuracy), nor has it been shown that one could construct generalized models applicable to a variety of books with reasonable results (above 90% character accuracy).

# 3 Diachronic Corpora and RIDGES

Our method is evaluated using a *diachronic* corpus because diachronic corpora with their extremely high variability are a good test case for the training and application of our OCR models. Many historical corpora are basically synchronic, covering a given linguistic period.[13] In addition to these synchronic corpora, there is a small number of diachronic corpora built specifically to do research on change phenomena. Over and above the variation that comes through different dialects, traditions, text types, etc. which is present in any corpus, there is the variation that is caused by linguistic and extralinguistic change. The construction of diachronic corpora is subject to all the problems and decisions faced in the construction of synchronic historical corpora, such as corpus design decisions in a situation where many text types or dialects are simply not available, issues of choosing the *original text* (e.g. manuscript/print or edition), questions of tokenization, normalization, or annotation, etc. (see Rissanen 1989, Lüdeling, Poschenrieder, and Faulstich (2004), Rissanen (2008), Archer et al. (2015), Gippert and Gehrke (2015), among many others). Linguistic changes that make the construction of diachronic corpora more difficult are, among many others, changes in spelling, changes in word formation, or changes in syntax such as changes in word order.

Extralinguistic changes that might prove problematic include the change of medium (manuscript, print), but also changes within a medium such as different scripts, changing conventions with respect to layout, abbreviations, inclusion of images, the development of technical equipment (i.e. the printing press) or paper quality (Eisenstein 1979; Weel 2011) as well as the development of science with its methods and conventions, the foundation of universities and the addition of new areas of research (W. P. Klein 1999).

[13]Basically, even the so-called *synchronic* corpora for older language stages are often diachronic, covering a long period of time; examples are the reference corpora in the DeutschDiachronDigital project, such as the Referenzkorpus Altdeutsch (http://www.deutschdiachrondigital.de/home/) or the Old and Middle English Corpora listed at http://users.ox.ac.uk/-stuart/english/med/corp.htm

Most existing high-quality historical (synchronic and diachronic) corpora are transcribed, collated, and corrected manually, sometimes by (offshore) double-keying transcription. If OCRed text is available for historical prints at all, the quality is typically too low for linguistic studies (Strange et al. 2014; Piotrowski 2012).

Manual transcription is time-consuming and expensive and it requires well-trained transcribers. It would therefore be very useful for historical studies to have accesss to high quality OCR.[14] The RIDGES corpus which we use in this article to test our methods is being constructed for research on the development of the scientific register in German (similar corpora are available for English, see e.g. the corpora constructed by the Varieng group in Helsinki[15]). It contains herbal texts from between the earliest printing to about 1900. Herbal texts are chosen because they are available throughout the history of German (see e.g. Riecke 2004; Gloning 2007) and are among the earliest scientific texts that are available in a vernacular language in Europe. These texts are often compilations or loose translations or transmissions from Latin herbal compendia (transmitting authoritative texts from famous physicians such as Dioscorides, Galen, Avicenna, or Paracelsus) and can be viewed as predecessors of scientific as well as popular texts about plants and illnesses (Habermann 2003; W. P. Klein 2011).

The current version of RIDGES (5.0)[16] contains excerpts (about 30 pages each) from 33 different books roughly spaced between 1487 and 1914. The originals are very different from each other in appearance. The earliest texts are mostly collections of *herbal monographs*, that is descriptions of a given herb with respect to its properties according to the theory of humoral pathology and its medical indications and ways of preparation for treatment. Starting in the 16th century, we increasingly find botanical facts. Later texts show greater variety: some are clearly botanical or medical, others are more in the realm of popular science. Because RIDGES is used for register studies (Biber and Conrad 2009), it is necessary to analyse and annotate many different properties (lexical, morphological, syntactic, structural, etc.) of each text. The corpus is therefore stored in a multilayer architecture; annotation layers can be added at any time (Krause and Zeldes 2016). The texts are transcribed diplomatically, normalized on several layers, and annotated deeply (some of the texts contain as many as 53 annotation layers).

[14] While our focus here is on linguistic research questions it is obvious and has been argued many times (see e.g., the DFG Practical Guidelines on Digitisation [02/13], p. 29; http://www.dfg.de/formulare/12_151/) that the better the OCR quality, the more useful the resulting text is for any kind of research. The re-usability of corpora has been an issue in many recent endeavours (cf. Lüdeling and Zeldes (2008), Geyken and Gloning (2015), among many others) but cannot be discussed here.

[15] http://www.helsinki.fi/varieng/about/index.html

[16] http://korpling.german.hu-berlin.de/ridges/index_en.html. The corpus is freely available under the CC-BY license. It can be downloaded in several formats as well as queried through the ANNIS search tool (Krause and Zeldes 2016). The corpus is still growing. Here we focus only on the first step in corpus compilation. More on meta-data, annotation, and analysis can be found in Odebrecht et al. (2017, in press).

For our present purpose the most important layer is the diplomatic transcription which serves as training material (ground truth) for the OCR engine (see below), and we want to briefly explain the decisions we took to prepare the diplomatic layer. For all the texts in RIDGES, scans are available from libraries (mostly from Bavarian State Library[17]) or from other sources such as Google Books.[18] Early texts are transcribed manually by students and collated and corrected by other students and researchers; for later texts it is sometimes faster to correct existing OCR results available from library digitization programs or Google Books. The transcription is highly diplomatic - differences between letter forms (like ſ and s) are preserved as well as abbreviation signs (e.g. *vñ* for modern German *und*, *modoꝝ* for *modorum* with Latin abbreviation ꝝ), or other special characters. Words broken up at a line break are also broken up in the transcription. Layout information (line breaks, page breaks, etc.), information about typefaces (Antiqua, broken fonts), and rendering (colours, spaced type, etc.) are provided in the annotation.

# 4 Training the OCR engine on historical printings

Our training procedure is based on the OCRopus software which was the first OCR engine with a recognition algorithm based on recurrent neural nets (RNNs) with long short-term memory (LSTM; Hochreiter and Schmidhuber 1997).[19] The LSTM architecture overcomes the problem of earlier neural networks to forget previously learned information and has proven to be very successful in pattern recognition tasks such as handwriting recognition (Graves et al. 2009), even in the context of medieval manuscripts (Fischer et al. (2009) got a word accuracy of 93% on isolated words). Models trained with lots of computer fonts led to excellent recognition results for 19th and 20th century printings of Antiqua and Fraktur (blackletter) typefaces with accuracies above 98.5% (Breuel et al. 2013). As mentioned in the introduction, for earlier printings it is necessary to train on real data. Images of printed lines have to be matched with their corresponding transcriptions (see Figure 2). Unicode code points for unusual historical characters are available due to the efforts of the Medieval Unicode Font Initiative.[20]

[17] http://www.digitale-sammlungen.de/index.html?c=digitale_sammlungen&l=en

[18] https://books.google.de/

[19] Tesseract in its upcoming version 4 also uses LSTM networks, but its training procedure is quite resource intensive and there is not yet any documentation available on how to train a recognition model on real printed images.

[20] http://folk.uib.no/hnooh/mufi/

The segmentation of the image of a printed page into single text lines is part of a preprocessing step that also involves other functions such as deskewing, border cropping, converting colored page images into binary or grayscale, and some denoising. While OCRopus has its own routines for doing that, they are rather basic and we used the open source program ScanTailor[21] for producing clean tif-images of text regions that were subsequently cut into text lines by the OCRopus subprogram *ocropus-gpageseg*.

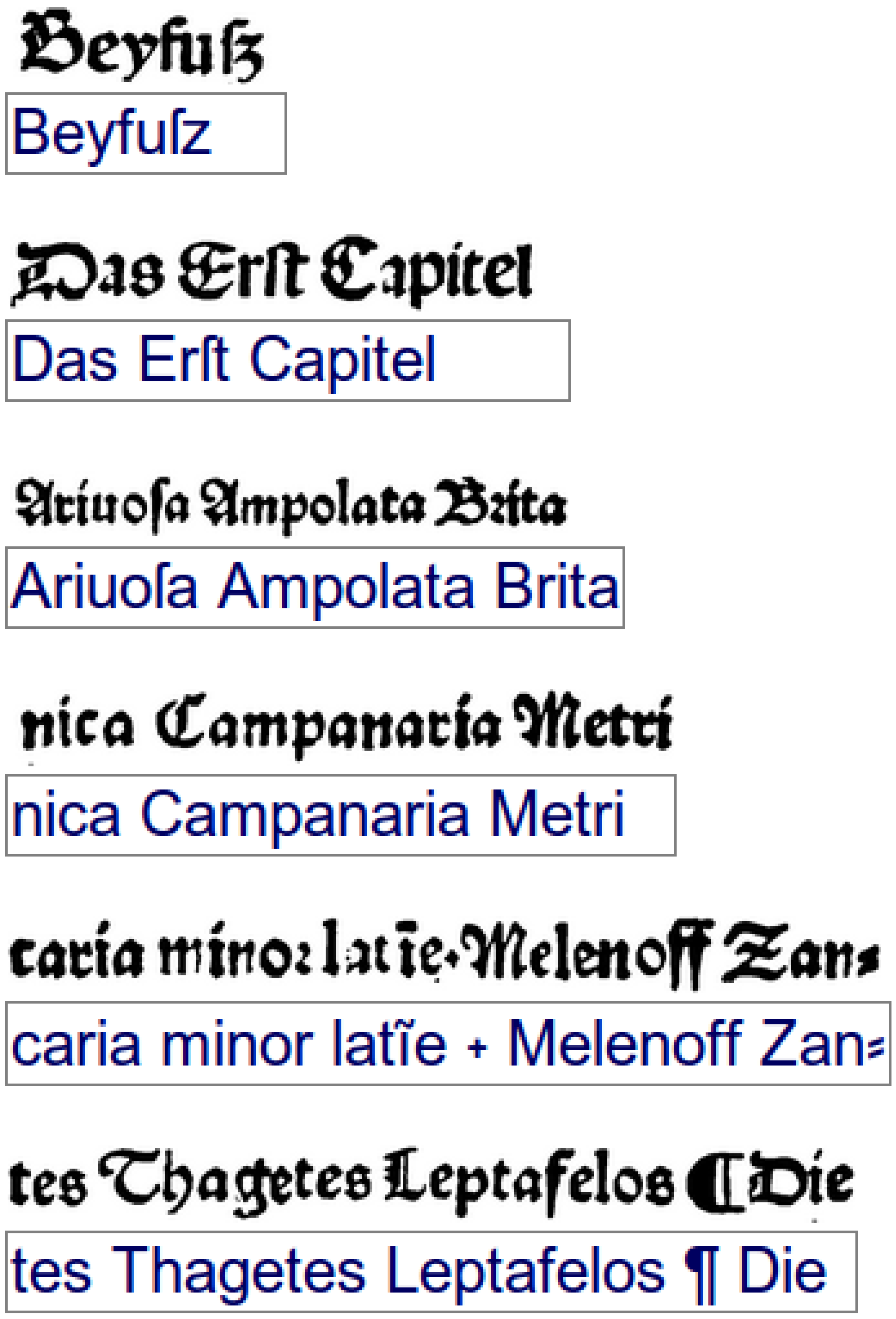

Figure 2: Pairs of line images and their associated transcriptions needed for training

During training, the OCR engine accesses these images-transcription pairs randomly and learns to associate the inputs (vertical stripes of pixel values) to outputs (characters). Be-cause the line image gets divided into many vertical slices each one pixel wide, there is no need to segment the line further into single glyphs as is usually done by traditional OCR. Instead, the engine learns to associate series of slices corresponding to single glyphs (including ligatures, digraphs etc. which are not easily segmented from neighboring glyphs) to characters or even character groups automatically (see Figure 3). The cumbersome reconstruction of fonts by cutting out images of the complete glyph repertoire necessary for the training of Tesseract is thus completely avoided.

[21]See http://scantailor.org.

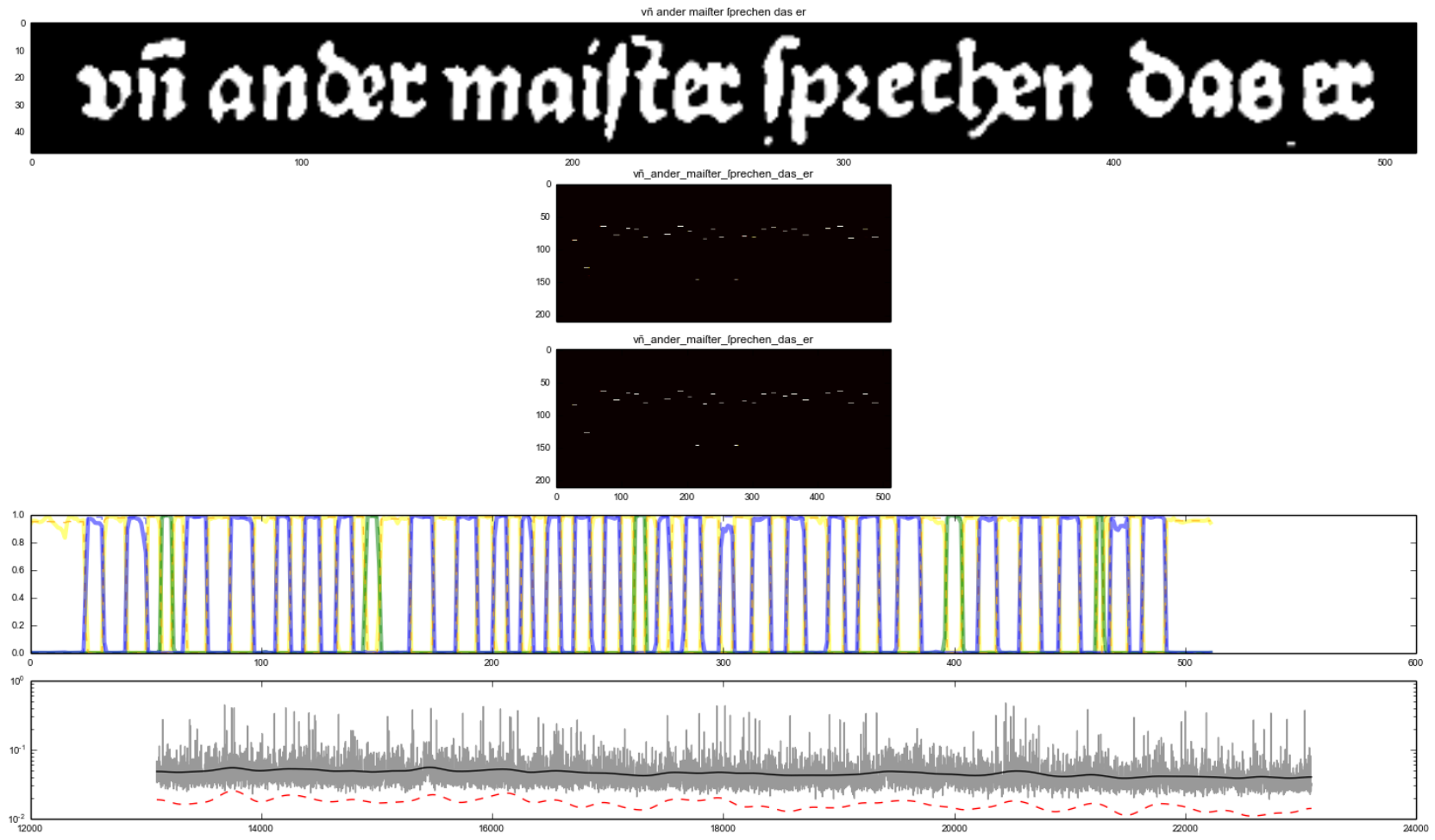

Figure 3: The training process in action: The neural network is presented with a printed text line together with its transcription (uppermost panel). The pixel coordinates at the x axis correspond to the vertical slices mentioned in the text. Groups of sequential slices corresponding to a glyph are learned automatically and labelled with the corresponding character from the transcription. The two black panels show the network response to the feeding of the line: the vertical axis represents the type case of the printer and enumerates single characters (about 200), while the horizontal axis again is the pixel width of the line. At each horizontal position, the number of the output character is highlighted. The next panel shows the confidence by which the network recognized the glyphs on a scale from 0 to 1. Blue rectangles correspond to characters, green rectangles to inter-word spaces. The last panel gives the prediction error over training history (over 23,000 lines have been seen at this point).

More detailed explanations of the inner workings of neural net training are given in Breuel et al. (2013), and a detailed tutorial on how to train models from the user perspective is available in Springmann (2015). Special recommendations for training models for incun-able printings are given in Springmann and Fink (2016).[22] Springmann, Fink, and Schulz (2016) show that a training set of just 100–200 randomly selected lines with their transcription often results in a model that is almost as good in its performance as more extensive training sets. In this study we used all the available material (i.e. the ca. 30 transcribed pages for each book) and split it into 90% for training and 10% for testing.

Every 1,000 learning steps (one step consists in seeing one image-transcription pair) an OCR model is saved to disk, and after having seen each pair about 30 times the training process is stopped. The series of models is then evaluated on the test set which has never been seen in training. Evaluation is done by a line-wise comparison of the string of symbols representing the ground truth and the OCR result, and the minimal edit (Levenshtein) distance between these two symbols strings is the number of character errors for this line. Complete error statistics are available from the OCRopus command *ocropus-errs*. The model with the least error on this test when comparing OCR result and ground truth is then selected as the best one representing the training effort for this book. An example of the resulting OCR text together with the corresponding page image is shown in Figure 4.

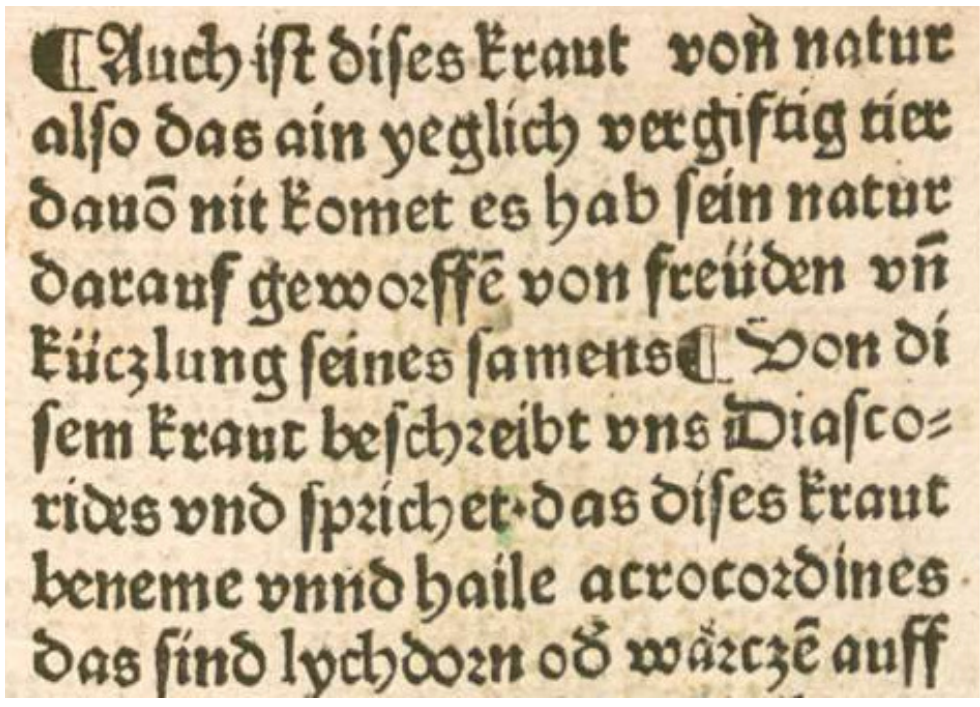

**¶ Auch ift difea kraut von natur**
**alfo das ain yeglich vergiftig tier**
**dauõ nit komet es hab fein natur**
**darauf geworffẽ von freüden vñ**
**küczlung feines famens¶ Von di**
**fem kraut befchreibt vns Diafco=**
**rides vnd fprichet + das difes kraut**
**beneme vnnd haile atrocordines**
**das find hychdorn od warczẽ auff**

Figure 4: Resulting OCR text from application of a trained model to a previously unseen page (1487 Garten der Gesunthait, errors marked in red).

# 5 OCR results from individually trained models

For the training of individual models we have selected the following 20 books[23] printed in broken typefaces typical of German historical printings. The printing dates cover a period of almost four centuries, from the incunable printing of *Garten der Gesunthait* in 1487 to *Deutsche Pflanzennamen* from 1870 (cf. Table 1).

[22] http://www.cis.lmu.de/ocrworkshop

[23] Sources are given at http://korpling.german.hu-berlin.de/ridges/download_v5.0_en.html.

Table 1: Books used for training and testing OCR models.

| Year | (Short) Title | Author | Label |
|---|---|---|---|
| 1487 | Garten der Gesunthait | Johannes von Cuba | 1487-G |
| 1532 | Artzney Buchlein der Kreutter | Johannes Tallat | 1532-A |
| 1532 | Contrafayt Kreüterbuch | Otto Brunfels | 1532-C |
| 1543 | New Kreüterbuch | Hieronymus Bock | 1543-N |
| 1557 | Wie sich meniglich | Adam von Bodenstein | 1557-W |
| 1588 | Paradeißgärtlein | Konrad Rosbach | 1588-P |
| 1603 | Alchymistische Practic | Andreas Libavius | 1603-A |
| 1609 | Hortulus Sanitatis | Castore Durante | 1609-H |
| 1609 | Kräutterbuch | Bartholomäus Carrichter | 1609-K |
| 1639 | Pflantz-Gart | Daniel Rhagor | 1639-P |
| 1652 | Wund-Artzney | Guilelmus Fabricius Hildanus | 1652-W |
| 1673 | Thesaurus Sanitatis | Adrian Nasser | 1673-T |
| 1675 | Curioser Botanicus | Anonymous | 1675-C |
| 1687 | Der Schweitzerische Botanicus | Timotheus von Roll | 1687-D |
| 1722 | Flora Saturnizans | Johann Friedrich Henckel | 1722-F |
| 1735 | Mysterium Sigillorvm | Israel Hiebner | 1735-M |
| 1764 | Einleitung zu der Kräuterkenntniß | Georg Christian Oeder | 1764-E |
| 1774 | Unterricht von der allgemeinen Kräuter- und Wurzeltrocknung | Johann Georg Eisen | 1774-U |
| 1828 | Die Eigenschaften aller Heilpflanzen | Anonymous | 1828-D |
| 1870 | Deutsche Pflanzennamen | Hermann Grassmann | 1870-D |

The resulting character and word accuracies for models trained on each book have been evaluated on unseen test pages and are shown in Figure 5. To get an idea of the variance of the measurement values we also indicated upper and lower limits of a 95% confidence interval for character accuracies calculated from the assumption that OCR recognition can be treatead as a Bernoulli experiment with the measured accuracy as the probability for correct recognition. For each book, the individual model is able to recognize printed characters with an accuracy of over 94%, sometimes even reaching 99%.

Word accuracies are also given, as they are important for search and indexing purposes (Tanner, Muñoz, and Ros 2009). We have to keep in mind, though, that early printings show a high variety of spellings and a high word accuracy does not necessarily mean that we find every instance of a correctly recognized word by searching for its modern equivalent. Evaluations of word accuracies were done with the UNLV/ISRI Analytic Tools for OCR Evaluation (Nartker, Rice, and Lumos 2005) adapted for UTF-8 by Nick White.[24] Compared to character accuracies, the word accuracies show a much wider variance, ranging from 76% to 97%. Except for two instances (1675-D and 1687-D), however, all models have word accuracies over 85%, twelve models over 90%, and five even over 95%, resulting in a mean word accuracy of 91%. This is in stark contrast to the newspaper projects reported in Tanner, Muñoz, and Ros (2009) with a mean word accuracy of 78% for the 19th Century Newspaper Project and 65% for the Burney Collection (17th and 18th century newspapers) and illustrates the progress in recognition that has become possible through Neural Network-based OCR models.

The wider variance for word accuracies can be explained by the statistics of single character errors. Assuming that we had 600 characters with 30 errors (95% character accuracy), and that the 600 characters consist of 100 words, the 30 character errors could be contained in just 5 words or in 30 words with resulting word accuracies ranging from 95% down to 70%. In the latter case, erroneous words just contain a single character error and a better model with half of the errors would therefore upgrade character accuracy from 95% to 97.5%, but word accuracy would get a much higher boost from 70% to 85%.

The most striking result reported in Figure 5 is the lack of any correlation between OCR accuracy and printing age: Models for even the earliest printings in our corpus show top performance when the printings were well preserved and good scans were available.

These results have been achieved just by recognizing glyph shapes, without employing any language model, lexicon or postcorrection. This shows that OCR quality does not deteriorate with earlier printing dates but instead depends on the quality of the printings in its currently preserved state and the scans. Lower accuracies point to instrinsic problems on book pages such as manual annotation, speckles, or low resolution (about 150 dpi), especially when downloaded from Google Books. Often a higher resolution of the same book was available from the Bavarian State Library (BSB).[25] Given good printings in well preserved states, 300 dpi color or gray-scale scans, and an accurate preprocessing of page images into single text lines (see the discussion in Section 7), it is possible to reach character accuracies above 98% in all periods of modern printing. The fact that commercial OCR applications yield increasingly worse results the older the printing just reflects that fact that earlier printings are increasingly different from the fonts the OCR engine has been trained on.

While the above results are very promising, they could only be reached by training an individual model on each book. Using RIDGES, we are in the favorable situation that the necessary ground truth needed for training is available. In general, however, no such transcriptions will exist and this may still prevent the application of OCR to large volumes of scans of historical books in an automatic manner. It is therefore of interest to see how well the existing models generalize to unseen books. To test how the individual models perform on the other books of our selection, we applied each model to each book. The resulting accuracies from this experiment are shown in Figure 6. The rows represent the books, whereas columns designate the models.

[24] Available from https://ancientgreekocr.org/

[25] E.g., a retraining on the BSB scan of the same exemplar of Bodenstein's 1557 book increased the accuracy from 95% to 99%.

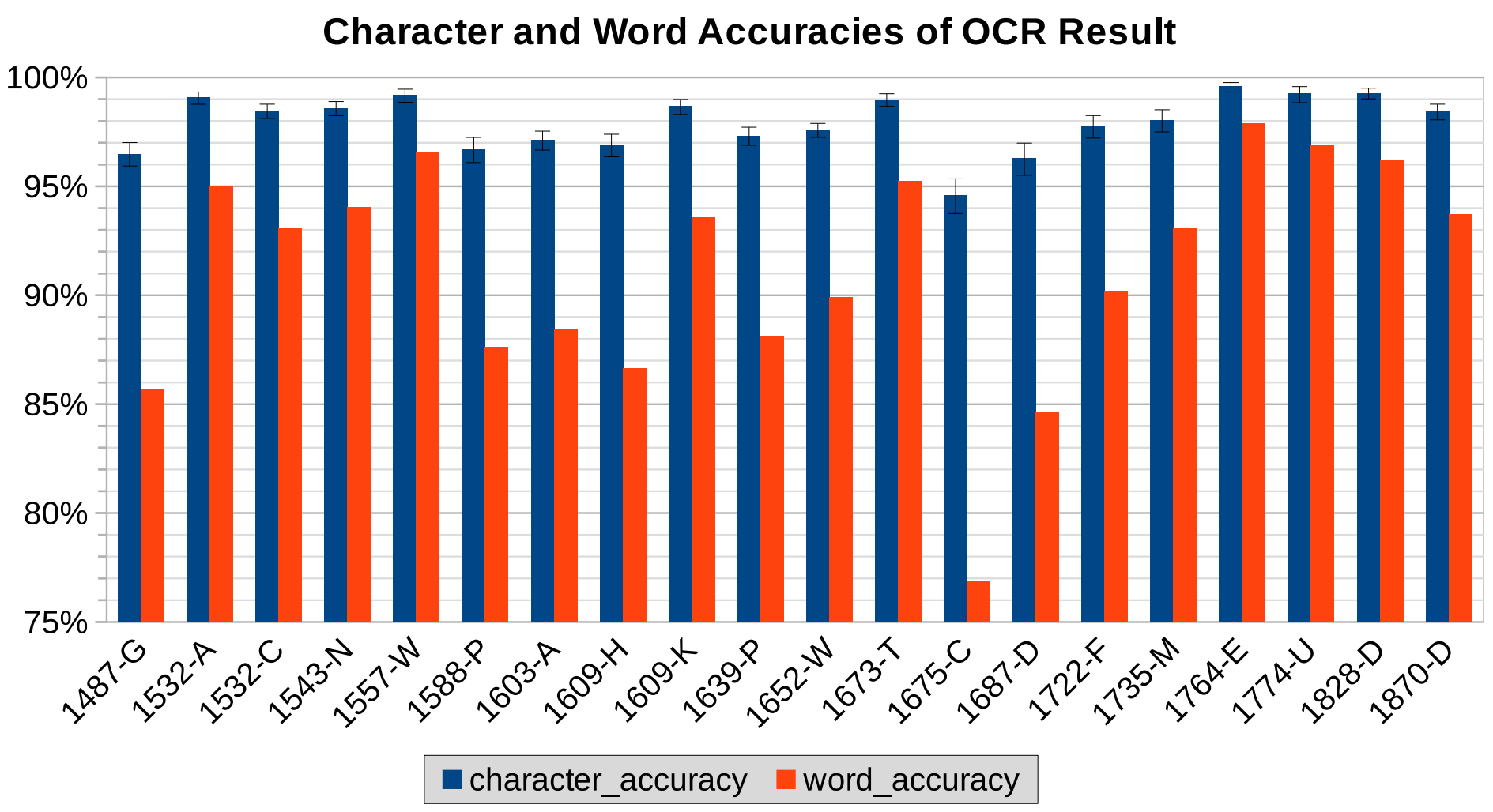

Figure 5: Character accuracies (blue columns) with 95% confidence intervals and word accuracies (red columns) for models trained on the individual books given in Table 1.

As expected, the best result for each book is provided by its own model. This is mostly true for the models as well, but in one case (1675-C) the trained model performs better on other books with a similar font (1764-E and 1774-U, although not as well as these books' models) than on its own book. The bad performance of 1675-C is due to the grayscale images lacking in contrast (see Figure 1). More interesting is the question whether individual models give satisfactory results on any other book. The model for 1675-C is an exception with accuracies over 90% for books printed between 1609 and 1828. If accuracies above 90% are acceptable, models trained on printings from the 17th century seem to be somewhat applicable to printings from the 17th and 18th century, whereas for later and especially earlier printings single models do not generalize well. In the next section, we will therefore explore whether mixed models trained on several books will better generalize to unseen books and offer a way to overcome the typography barrier.

| | 1487-G | 1532-A | 1532-C | 1543-N | 1557-W | 1588-P | 1603-A | 1609-H | 1609-K | 1639-P | 1652-W | 1673-T | 1675-C | 1687-D | 1722-F | 1735-M | 1764-E | 1774-U | 1828-D | 1870-D |
|---|---|---|---|---|---|---|---|---|---|---|---|---|---|---|---|---|---|---|---|---|
| **1487-G** | **96.5** | 77.0 | 74.7 | 75.4 | 79.3 | 72.4 | 74.1 | 68.1 | 72.6 | 73.8 | 71.0 | 70.4 | 77.5 | 64.6 | 72.6 | 71.3 | 66.6 | 72.2 | 64.2 | 54.7 |
| **1532-A** | 84.3 | **99.1** | 85.4 | 85.7 | 90.5 | 84.8 | 90.5 | 87.2 | 89.4 | 83.5 | 87.2 | 81.9 | 90.6 | 79.7 | 74.1 | 84.2 | 67.9 | 75.8 | 70.3 | 59.2 |
| **1532-C** | 80.0 | 74.9 | **98.5** | 84.5 | 84.4 | 72.9 | 80.9 | 74.5 | 67.7 | 77.8 | 63.8 | 67.0 | 80.7 | 60.1 | 67.5 | 66.3 | 63.6 | 67.6 | 56.2 | 46.4 |
| **1543-N** | 89.9 | 88.6 | 91.0 | **98.6** | 91.6 | 85.2 | 86.9 | 85.0 | 86.1 | 85.5 | 80.5 | 81.5 | 88.9 | 74.9 | 80.0 | 84.3 | 73.8 | 75.7 | 71.1 | 60.8 |
| **1557-W** | 90.0 | 84.8 | 87.6 | 84.0 | **99.2** | 82.1 | 83.6 | 79.8 | 76.7 | 84.1 | 83.4 | 70.2 | 89.7 | 69.1 | 73.4 | 78.9 | 66.9 | 79.5 | 76.4 | 63.8 |
| **1588-P** | 69.2 | 71.2 | 66.9 | 66.8 | 72.2 | **96.7** | 86.6 | 86.2 | 85.1 | 88.4 | 90.3 | 84.8 | 88.6 | 82.1 | 76.4 | 79.1 | 72.3 | 74.9 | 70.0 | 62.3 |
| **1603-A** | 78.4 | 81.9 | 79.7 | 78.5 | 78.5 | 89.0 | **97.1** | **95.7** | 91.4 | 90.0 | 83.8 | 87.9 | 87.5 | 84.6 | 85.7 | 84.6 | 76.3 | 76.6 | 64.3 | 63.1 |
| **1609-H** | 67.7 | 72.8 | 72.4 | 69.3 | 68.8 | 86.4 | 93.6 | **96.9** | 87.8 | 84.3 | 80.0 | 81.5 | 82.9 | 78.2 | 76.5 | 76.9 | 65.3 | 66.9 | 59.6 | 58.9 |
| **1609-K** | 83.1 | 83.4 | 81.6 | 82.6 | 83.3 | 93.9 | **97.0** | **96.2** | **98.7** | 92.7 | 92.1 | 90.9 | 93.3 | 91.5 | 84.7 | 88.2 | 80.3 | 82.5 | 76.7 | 68.0 |
| **1639-P** | 79.7 | 80.1 | 77.7 | 79.3 | 82.0 | 91.8 | 92.6 | 91.7 | 91.0 | **97.3** | 94.5 | 89.3 | 93.6 | 86.7 | 86.2 | 86.9 | 81.1 | 86.5 | 75.8 | 70.1 |
| **1652-W** | 71.5 | 77.1 | 71.4 | 61.4 | 76.7 | 91.6 | 89.0 | 85.8 | 85.8 | 92.4 | **97.6** | 87.8 | 92.0 | 86.8 | 82.7 | 84.8 | 78.8 | 83.0 | 72.8 | 66.1 |
| **1673-T** | 73.0 | 79.1 | 70.3 | 69.0 | 77.3 | 88.8 | 91.8 | 88.7 | 90.6 | 90.3 | 91.1 | **99.0** | 93.5 | 90.6 | 87.8 | 88.2 | 86.3 | 83.9 | 78.3 | 70.3 |
| **1675-C** | 72.0 | 72.6 | 73.3 | 76.3 | 75.8 | 88.5 | 82.7 | 85.3 | 84.9 | 91.7 | 89.1 | 82.4 | 94.6 | 80.8 | 78.7 | 80.9 | 76.8 | 79.4 | 73.0 | 66.2 |
| **1687-D** | 74.2 | 76.7 | 63.7 | 64.0 | 68.1 | 82.2 | 89.3 | 87.0 | 88.7 | 87.6 | 89.5 | 90.3 | 94.2 | **96.3** | 86.6 | 84.7 | 84.5 | 83.7 | 77.5 | 69.2 |
| **1722-F** | 75.8 | 71.5 | 70.5 | 72.2 | 73.2 | 81.4 | 88.5 | 84.7 | 84.7 | 89.3 | 83.5 | 87.3 | 92.2 | 84.7 | **97.8** | 91.6 | 87.5 | 86.9 | 77.0 | 73.0 |
| **1735-M** | 79.0 | 80.1 | 77.8 | 81.0 | 82.5 | 85.1 | 90.8 | 86.1 | 87.6 | 91.6 | 87.3 | 90.1 | 92.0 | 86.8 | 94.7 | **98.1** | 90.8 | 91.5 | 86.9 | 85.1 |
| **1764-E** | 82.7 | 78.2 | 73.8 | 70.3 | 78.2 | 91.3 | 88.8 | 85.7 | 88.4 | 93.6 | 92.5 | 95.0 | **97.2** | 91.1 | **95.6** | **95.3** | **99.6** | **96.2** | 93.0 | 88.4 |
| **1774-U** | 81.6 | 80.6 | 79.9 | 76.3 | 84.6 | 92.7 | 92.6 | 90.5 | 90.3 | **95.8** | **95.5** | 93.0 | **96.5** | 91.2 | 94.4 | **95.5** | **96.4** | **99.3** | 94.3 | 87.2 |
| **1828-D** | 75.2 | 77.0 | 77.3 | 67.3 | 78.6 | 86.1 | 84.8 | 82.3 | 84.7 | 89.7 | 89.2 | 87.6 | 93.5 | 83.1 | 88.0 | 90.7 | 93.9 | 92.4 | **99.3** | 93.5 |
| **1870-D** | 71.3 | 71.6 | 69.2 | 65.6 | 69.9 | 81.3 | 80.4 | 80.1 | 79.8 | 84.9 | 82.3 | 84.5 | 87.4 | 81.3 | 86.1 | 84.2 | 86.6 | 84.5 | 88.2 | **98.4** |

Figure 6: Cross evaluation of each model (column) on each book (row). Character accuracies are color coded in the ranges greater than 95% (green), between 90 and 95% (yellow) and below 90% (gray).

# 6 Construction of mixed models

We trained two mixed models on a breakdown of our corpus into two parts by choosing every other book as part 1 and the rest as part 2 (Figure 7). Each part covers about 400 years of printing history so that we can test how well a mixed model trained from a sample of books over a wide range of printing dates generalizes to other unseen books dating from the same period.

The result of applying each of the two mixed models to the books from the other subset is shown in Figure 7 (yellows columns) together with the accuracies of individual models (blue columns). The results are given in terms of both character accuracy (upper panel) and word accuracy (lower panel).

As may be expected, mixed models give a lower accuracy compared to individual models (with the exception of 1774-U), because mixed models have not been trained on books of the other part to which they have been applied. But except for the two earliest printings and in contrast to Figure 6, both mixed models now consistently reach over 90% character accuracy on books which did not contribute to the training pool. In one case (1774-U) the mixed model shows a slightly better performance than the individual model. This book is similar enough in its typography to the books of the other part on which the mixed model has been trained, so that recognition is improved. The incunable printing of 1487 gets the worst recognition from the mixed model because it is unlike in its typography to any other book.

Comparing the difference in character and word accuracies, we see the effect that the improved character accuracies that are obtained from going from a mixed model to an individual model also improve the word accuracies, but by a larger absolute margin (roughly, halving the character errors also halves the word errors; see Table 2).

Table 2: Mean accuracies for individual and mixed models

| mean accuracy | Part 1 | Part 2 |
|---|---|---|
| *character accuracy* | | |
| individual models | 98.07% | 97.94% |
| mixed model | 95.81% | 94.27% |
| *word accuracy* | | |
| individual models | 91.53% | 90.71% |
| mixed model | 83.20% | 76.09% |

The remaining OCR errors in both the mixed and individual model recognition consist in equal numbers of substitution errors (one character gets confused with another, often similar looking character, such as *e* and *c*) and deletion or insertion errors (characters are either not recognized at all or spurious characters are inserted where none were printed). By far the most frequent deleted or inserted character is the blank, which when inserted leads to broken-up words into two or more fragments, or if deleted, leads to merged words where two or more printed words are recognized as a single word. This again underscores the necessity to train a recognition model on real printings, as only in this way a proper word distance model can be learned, and even then the peculiarities of early-modern printing with its emphasis on justified printing lines lead to relatively frequent errors involving the non-character of blanks.

The training of mixed models does therefore seem to be a way to overcome the typography barrier and it can provide models which generalize well over a wide range of books. The resulting OCR text can at least be taken as a first approximation and better models can be trained from an error corrected version of its recognized text. The same conclusion has been reached in a study on twelve Latin books printed from 1471 to 1686 in Antiqua fonts in Springmann, Fink, and Schulz (2016), where a method is given to construct better individual models with minimal manual effort starting from a mixed model.

# 7 Discussion

For the practical purpose of historical corpus construction, it is very encouraging to see that we can train OCR models that work very well for even the oldest printings. However, this success comes at the price of necessary training for each new font. The peculiarities of early typesetting requires that we train on real data and we must therefore manually prepare the ground truth for a couple of pages. The resulting trained model will be applicable to all works printed with the same font, which for early printings means the same font of the same typographer's office which may have been used for a whole series of works. The possibility to train mixed models covering a range of historical fonts and periods is a big step forward to convert historical printings to electronic text without having to train an individual model for each text. In the absence of ground truth, one can start with the output of a mixed model and only correct it if needed, so that manual transcription is reduced to a minimum of about 100 to 200 text lines (Springmann, Fink, and Schulz 2016).

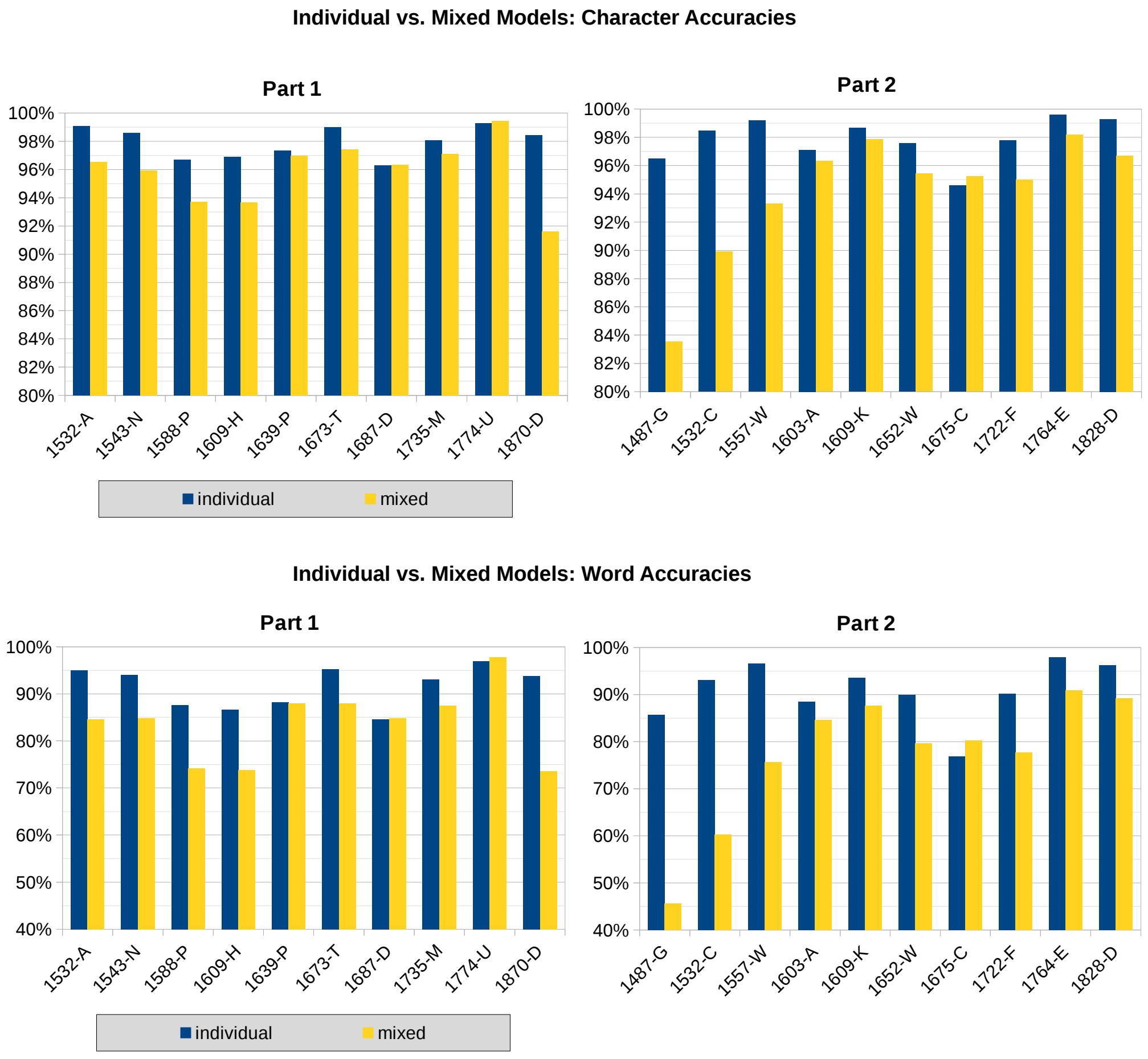

Figure 7: Accuracies of OCR results for individual models (blue) and for a mixed model trained on the other part of the corpus (yellow).

We want to stress that the usefulness of OCR results for research purposes depends not only on recognition quality, but also on further preprocessing (page segmentation, binarization etc.) and postprocessing steps (error correction, normalization, annotation) which we will discuss in turn.

Because text recognition happens on single printed lines, we need to segment printed pages into those lines. This is not a trivial task, as many historical printings abound with decorative initials, floral decorations, marginal annotations (printed and manual), and image drawings (see Figure 1). Anything that goes wrong at this preprocessing stage will lead to bad recognition results later. Because the preprocessing routines of the OCR engines are often not able to solve this difficult task in a satisfactory automatic manner there is a need for specific tools which could at least assist a human to semi-automatically segment the page images. ScanTailor is too restricted when the layout becomes too involved with text columns, headings spanning the columns, marginal and footnotes etc. To fill this gap, a new open source tool has been made available (LAREX: Reul, Springmann, and Puppe 2017) that allows the user to quickly interact with the layout, define some general parameters and have a complete document segmented automatically, including the reading order. Errors occuring in this automatic process can afterwards be amended manually. In a case study treating an incunable printing consisting of almost 800 pages, the complete document could be segmented into text zones by an experienced user in less than six hours (Reul, Dittrich, and Gruner 2017).

On the postprocessing side, we first have to deal with the accuracy of the OCR result. Whether the uncorrected OCR output is sufficient for research or not depends on the specific research interests of a scholar. 95% character accuracy (or 90% word accuracy) may be enough for searching within a work or a corpus, especially if the search is spelling tolerant and if the results are simultaneously highlighted in the scan. Getting a high percentage of possible hits *(high recall)* is more important than getting a low number of false hits *(high precision)*, because false hits can easily be discarded by looking at the scan. Any true hit will provide the researcher with a piece of evidence he is looking for. For other interests a higher accuracy might be needed and error correction will be necessary. The efficiency of correction will be greatly helped by a tool that presents whole series of statistically induced errors which the user can inspect, select and correct in a single step. The more common errors can then be corrected with little effort, boosting accuracy and leaving only rare errors for which one still would need to go over the complete text if needed. Such a postcorrection method enabling batch correction of common errors has been developed at the Center for Information and Language Processing of LMU under the name of PoCoTo (Vobl et al. 2014) and is available under an open-source licence.[26] Furthermore, the tool is able to distinguish between historical spellings and real OCR errors and offers only the latter category for correction. It can import OCR results from ABBYY Finereader (XML), Tesseract (hOCR), and OCRopus (hOCR) and export the corrected text as pure text, XML, hOCR, or TEI.

But even if we had 100% correct transcriptions from corrected OCR recognition, we would still be hampered by the fact that there is no standardized spelling for historical texts which results in a high degree of variation (the same word can be spelled in five different ways by the same author in the same text). Often, but certainly not always, there is an etymologically related modern word. Depending on the research question, there might be different ways of dealing with meaning change or etymological replacement. Searching then becomes problematic as searches are usually done by entering search terms in modern spelling. One needs to make sure that all tokens with their various historical spellings are found. How this can best be achieved with a combination of modern and historical lexica and a set of historical patterns by which modern and historical wordforms may get connected has been described by Gotscharek et al. (2011), Ernst-Gerlach and Fuhr (2006), and Ernst-Gerlach and Fuhr (2007).

The current paper only considers the production of OCR recognized text, but for corpus construction one would also like to have methods for automatic normalization of historical texts, so that the expansion of modern wordforms to multiple possible historical wordforms becomes unnecessary, and which would also allow further treatment of corpus texts such as part-of-speech tagging and lemmatization. Work in this direction has been done by Pilz et al. (2008), Baron, Rayson, and Archer (2009), Bollmann, Petran, and Dipper (2011), and Jurish (2013). The RIDGES corpus contains a special annotation layer with a manually normalized token for each historical token. Normalized layers (which in multi-layer architectures can be provided in addition to the diplomatic layers) allow searches across different spellings (stemming from different authors and times) and can be used as input for automatic tools. Odebrecht et al. (2017, in press) illustrate this using several test cases. Weiß and Schnelle (2016) sketch how the normalized layer of RIDGES can be used for automatic parsing.

Finally, one has too keep in mind that residual recognition errors present in any OCR

[26] https://github.com/cisocrgroup/PoCoTo

corpus may exert a bias on research (see Traub, Ossenbruggen, and Hardman 2015), so that at least an indication of OCR quality in the absence of any ground truth would be highly desirable to give researchers an idea of what quality they are dealing with. Some work in this direction is reported in Springmann, Fink, and Schulz (2016).

To summarize, the current main obstacles for a machine-assisted construction of historical corpora are the problems of automatic segmentation of page images into text and non-text zones, the regnition of historical fonts, and the postprocessing problems of error correction and the normalization of historical spellings. The better we solve the recognition problem by means of trained historical OCR and mixed models, the more prominent the preprocessing (segmentation) and postprocessing (error correction, normalization) problems will become.

# 8 Summary and Conclusion

We have shown that the new trainable OCR method based upon LSTM RNNs can provide high recognition accuracies (character accuracies from 95% to over 99%, word accuracies from 76% to 97%) for images of early printings regardless of their age, provided good scans are available. An OCR model trained on the diplomatic transcription of some pages can be used for the recognition of the rest of the book and leads to electronic text whose quality may already be sufficient for many research questions. The availability of generalized models trained on a mixture of fonts would further ease the use of OCR on historical documents, as the output of these mixed models can be taken as a first approximation to a true representation of the text and may be corrected (and better models trained on it) if needed. The models trained on RIDGES have been used to recognize whole books and thus led to an expanded, if somewhat inaccurate, electronic corpus. Both this corpus (named RIDGES-OCR) and the above mentioned mixed models are available under CC-BY.[27] As more and more diplomatically transcribed texts (either manually transcribed or recognised by OCR methods) become available, their usability as a corpus for research purposes will depend on automatic means to normalize historical spelling (even in the presence of OCR errors) enabling both search and indexing as well as further analysis such as part-of-speech tagging and lemmatization.[28]

[27] https://www.linguistik.hu-berlin.de/en/institut-en/professuren-en/korpuslinguistik/research/ridges-projekt/ocr?set_language=en

[28] It would be very beneficial if a consortium of libraries would set up a central repository for trained models that would otherwise end up on the hard disks of individual users and not be available for others who would unnecessarily need to train the same model again. Even better, the complete training procedure could be made available via a web interface which allows to transcribe some pages as ground truth, train a model (possibly through several refinement steps as explained above) and at the end provide the user with the recognized text in some marked-up format as well as adding the model and ground truth to the repository. That way both the individual user and the community at large would benefit from an ever-increasing amount

This work was partially funded by Deutsche Forschungsgemeinschaft (DFG) under grant no. LU 856/7-1.

of ground truth, corresponding image data and trained models.